# A Non-Parametric Bayesian Method for Inferring Hidden Causes

**Frank Wood**
Computer Science
Brown University
Providence, RI 02912
fwood@cs.brown.edu

**Thomas L. Griffiths**
Cognitive and Linguistic Sciences
Brown University
Providence, RI 02912
Tom_Griffiths@brown.edu

**Zoubin Ghahramani**
Department of Engineering
University of Cambridge
Cambridge CB2 1PZ, UK
zoubin@eng.cam.ac.uk

## Abstract

We present a non-parametric Bayesian approach to structure learning with hidden causes. Previous Bayesian treatments of this problem define a prior over the number of hidden causes and use algorithms such as reversible jump Markov chain Monte Carlo to move between solutions. In contrast, we assume that the number of hidden causes is unbounded, but only a finite number influence observable variables. This makes it possible to use a Gibbs sampler to approximate the distribution over causal structures. We evaluate the performance of both approaches in discovering hidden causes in simulated data, and use our non-parametric approach to discover hidden causes in a real medical dataset.

## 1 Introduction

A variety of methods from Bayesian statistics have been applied to the problem of learning the dependencies among a set of observed variables [1, 2]. However, in many settings, the dependencies of interest are not those that exist among the observed variables, but those that are produced by "hidden causes". For example, in medicine, the symptoms of patients are explained as the result of diseases that are not themselves directly observable – an assumption that is embodied in graphical models for medical diagnosis, such as QMR-DT [3]. Here we consider how Bayesian methods can be used to infer both the existence of such hidden causes and how they influence observed variables. In our medical example, this would mean discovering diseases from the symptoms of patients.

Learning the structure of graphical models containing hidden causes presents a significant challenge, since the number of hidden causes is unknown and potentially unbounded. Researchers have explored several approaches to this problem. One approach uses statistical criteria to identify when hidden causes might be present (e.g., [4]). While these algorithms are effective, they do not reflect our aim of developing a Bayesian approach to solving this problem. Other more closely related work defines a prior over the number of hidden causes, and uses "reversible jump" Markov chain Monte Carlo (RJMCMC) [5] algorithms to move between structures with different numbers of hidden causes (e.g., [6, 7]). These methods satisfy our desire for a Bayesian solution, but designing well-mixing RJMCMC algorithms can be difficult.

Previous Bayesian approaches to inferring hidden causal structure assume that the number of hidden causes is finite. In many cases, it is more accurate to assume instead that the number of hidden causes is infinite. Rather than seeking to determine the number of hidden causes, we instead seek to find and count the finite subset of these hidden causes that manifest in a particular finite dataset. This perspective on the dimensionality of models is common in non-parametric Bayesian statistics. For example, Dirichlet process mixture models assume that data come from a potentially infinite number of clusters, of which only a finite subset are observed [8, 9, 10]. However, non-parametric Bayesian methods have not previously been applied to the problem of learning causal structure from data.

In this paper we develop a non-parametric Bayesian approach to structure learning with an unbounded number of hidden causes. Specifically, we define a prior over causal structures using the Indian buffet process (IBP) [11], a distribution over infinite binary matrices. Using the properties of the IBP, we derive a Gibbs sampling algorithm that can be used to sample from the posterior distribution over causal structures. We compare this approach to standard RJMCMC methods, and use it to infer the hidden causes behind the symptoms of stroke patients.

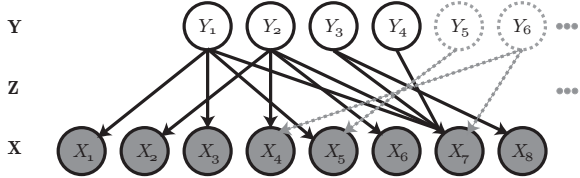

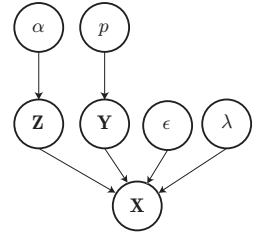

Figure 1: A hypothetical Bayesian network connecting hidden causes $Y_1, \ldots, Y_K$ to observed variables $X_1, \ldots, X_N$. We consider the case where the number of hidden causes, $K$, is unbounded. The state of the hidden causes, the observed variables, and the dependencies between them can all be summarized using binary matrices, being $\mathbf{Y}$, $\mathbf{X}$, and $\mathbf{Z}$ respectively.

## 2  Learning Hidden Causal Structure

Assume we have observed $T$ instances (trials) of a set of $N$ binary variables. Using $x_{i,t}$ to denote the value of the $i^{\text{th}}$ observed variable on the $t^{\text{th}}$ trial, we can summarize these observations with an $N \times T$ binary matrix $\mathbf{X}$. A standard structure learning task would be to learn a Bayesian network representation of the dependencies among the variables $X_1, \ldots, X_N$. These dependencies can be expressed using the $N \times N$ adjacency matrix $\mathbf{A}$ of a directed graph in which the nodes correspond to the $N$ variables, with $a_{ij} = 1$ if an edge exists from node $j$ to node $i$, and 0 otherwise.

To extend this problem to include hidden causes, assume that there are a further $K$ binary variables which are never observed. Using $y_{k,t}$ to denote the value of the $k^{\text{th}}$ hidden cause on the $t^{\text{th}}$ trial, we can summarize the state of the hidden causes with a $K \times T$ binary matrix $\mathbf{Y}$. If we let $z_{i,k} = 1$ if an edge exists from node $k$ to node $i$, and 0 otherwise, we can represent the dependencies between hidden causes and observable variables with the $N \times K$ binary matrix $\mathbf{Z}$. The full adjacency matrix $\mathbf{G}$ for the Bayesian network defined on the variables $X_1, \ldots, X_N$ and $Y_1, \ldots, Y_K$ is

$$\mathbf{G} = \begin{bmatrix} \mathbf{A} & \mathbf{Z} \\ \mathbf{0} & \mathbf{0} \end{bmatrix} \quad (1)$$

where $\mathbf{0}$ denotes a matrix of zeros of the appropriate size to make $\mathbf{G}$ a square $(N+K) \times (N+K)$ matrix.

Our focus on this paper will be on learning $\mathbf{Z}$, as Bayesian methods for learning $\mathbf{A}$ (e.g., [1]) can easily be combined with the methods described here to infer $\mathbf{G}$ as a whole. Our problem, then, reduces to learning the structure of a bipartite graph (see Fig. 1).

## 3  Generative Model

Our goal is to infer the dependencies between the hidden causes and the observed variables, $\mathbf{Z}$, and the values of those hidden causes on each trial, $\mathbf{Y}$, from the values of the observed variables, $\mathbf{X}$. If we define a generative model specifying a distribution over $\mathbf{X}$, $\mathbf{Y}$, and $\mathbf{Z}$, we can compute the posterior distribution over $\mathbf{Z}$ and $\mathbf{Y}$ given $\mathbf{X}$ using Bayes' rule

$$P(\mathbf{Z}, \mathbf{Y}|\mathbf{X}) = \frac{P(\mathbf{X}|\mathbf{Y},\mathbf{Z})P(\mathbf{Y})P(\mathbf{Z})}{\sum_{\mathbf{Y},\mathbf{Z}} P(\mathbf{X}|\mathbf{Y},\mathbf{Z})P(\mathbf{Y})P(\mathbf{Z})} \quad (2)$$

Figure 2: Graphical model illustrating dependencies among variables in the generative model.

where we make the independence assumptions illustrated in Figure 2. We will start by assuming $K$ is finite, and then consider the case where $K \to \infty$.

### 3.1  A Finite Model

We assume that the entries of $\mathbf{X}$ are conditionally independent given $\mathbf{Z}$ and $\mathbf{Y}$, and are generated from a *noisy-OR* [12] distribution, with

$$P(x_{i,t} = 1|\mathbf{Z},\mathbf{Y}) = 1 - (1-\lambda)^{\mathbf{z}_{i,:}\mathbf{y}_{:,t}}(1-\epsilon) \quad (3)$$

where $\mathbf{z}_{i,:}$ is the $i^{\text{th}}$ row of $\mathbf{Z}$, $\mathbf{y}_{:,t}$ is the $t^{\text{th}}$ column of $\mathbf{Y}$, and $\mathbf{z}_{i,:}\mathbf{y}_{:,t} = \sum_{k=1}^{K} z_{i,k} y_{k,t}$. The baseline probability that $x_{i,t} = 1$ is $\epsilon$, and $\lambda$ is the probability with which any of the hidden causes is effective. This model makes sense in applications where many causes can elicit an effect, and the likelihood of observing an effect is increased by the number of hidden causes that are active. The medical diagnosis application we consider later in the paper fits this description well.

We assume that the entries of $\mathbf{Y}$ are generated independently from a Bernoulli($p$) distribution,

$$P(\mathbf{Y}) = \prod_{k,t} p^{y_{k,t}}(1-p)^{1-y_{k,t}} \quad (4)$$

where the product ranges over all values of $k$ from 1 to $K$ and all values of $t$ from 1 to $T$. This model makes only the very general assumption that the baseline "prevalence" of all hidden causes is roughly the same. This assumption may not be appropriate for some applications, and can be relaxed if necessary.

In specifying a distribution on $\mathbf{Z}$, our goal is to generate matrices that allow multiple hidden causes to affect the same observed variable. This characteristic is desirable in many settings, and is exemplified by the case of medical diagnosis, where multiple diseases can cause the same symptom. A simple process for generating such a $\mathbf{Z}$ would be to assume that each hidden cause $k$ (corresponding to a column of the matrix) is associated with a parameter $\theta_k$, and then sample the values of $z_{ik}$ from a Bernoulli($\theta_k$) distribution for $i$ ranging from 1 to $N$. If we make the further assumption that each $\theta_k$ is generated from a Beta($\frac{\alpha}{K}$, 1) distribution and integrate out the $\theta_k$, the probability of $\mathbf{Z}$ is

$$P(\mathbf{Z}) = \prod_{k=1}^{K} \frac{\frac{\alpha}{K}\Gamma(m_k + \frac{\alpha}{K})\Gamma(N - m_k + 1)}{\Gamma(N + 1 + \frac{\alpha}{K})} \quad (5)$$

where $\Gamma(\cdot)$ is the generalized factorial function, with $\Gamma(x) = (x-1)\Gamma(x-1)$, and $m_k = \sum_i z_{i,k}$.

### 3.2 Taking the Infinite Limit

In non-parametric Bayesian statistics, it is common to define models with unbounded dimensionality by taking the infinite limit of models with finite dimensionality [9, 10]. In this spirit, we can consider what happens to the model defined above as $K \to \infty$. The distribution on $\mathbf{X}$ remains well-defined, and the only values of $\mathbf{Y}$ with which we need be concerned are those in rows that correspond to columns of $\mathbf{Z}$ for which $m_k > 0$. Thus, we need only consider what happens to Eqn. 5 as $K \to \infty$. If we attend only to the columns for which $m_k > 0$ and define a scheme for ordering those columns, we obtain the distribution

$$P(\mathbf{Z}) = \frac{\alpha^{K_+}}{\prod_{h=1}^{2^N-1} K_h!} \exp\{-\alpha H_N\} \prod_{k=1}^{K_+} \frac{(N - m_k)!(m_k - 1)!}{N!}$$

where $K_+$ is the number of columns for which $m_k > 0$, $K_h$ is the number of columns whose entries correspond to the binary number $h$, and $H_N = \sum_{i=1}^{N} \frac{1}{i}$ [11].

This distribution can be shown to result from the *Indian buffet process*, defined in terms of a sequence of $N$ customers entering a restaurant and choosing from an infinite array of dishes (corresponding to columns of $\mathbf{Z}$). The first customer tries the first Poisson($\alpha$) dishes (placing 1s in the appropriate columns). The remaining customers then enter one by one and pick previously sampled dishes with probability $\frac{m_{-i,k}}{i}$, where $m_{-i,k}$ is the number of customers who have already chosen the $k^\text{th}$ dish. After trying the shared dishes, each customer also then tries the next Poisson($\frac{\alpha}{i}$) new dishes. The distribution that results from this process is *exchangeable*: the probability of each binary matrix $\mathbf{Z}$ is unaffected by the order of the customers.

## 4 Inference Algorithms

Having defined a generative model, we can use Bayesian inference to infer $\mathbf{Z}$ and $\mathbf{Y}$ from $\mathbf{X}$. However, since the denominator of Equation 2 is an intractable sum, an approximate inference algorithm must be used. We present two such algorithms: a RJMCMC algorithm for the model with finite but unknown dimensionality and a Gibbs sampler for the model with an unbounded number of hidden causes.

### 4.1 Reversible Jump MCMC

Reversible jump MCMC is a variant of the Metropolis-Hastings algorithm that allows moves between models of different dimensionality [5]. The central idea is to augment a sampler for a finite model with a "dimension-shifting" move. In our case, a standard "birth/death" proposal would be of the following form: pick a single hidden cause (column $k$ of $\mathbf{Z}$) and check the number of incident edges $m_k$. If $m_k = 0$, then remove that cause and decrement $K$. If $m_k > 0$, then add a new cause with no links, and generate the new values of $\mathbf{Y}$ that will correspond to it according to the prior. Letting $\xi$ denote the values of $\mathbf{Z}$, $\mathbf{Y}$, and $K$, the move is accepted with probability

$$A(\xi^*, \xi) = \min\left[1, \frac{P(\mathbf{X}, \xi^*)}{P(\mathbf{X}, \xi)} \frac{Q(\xi|\xi^*)}{Q(\xi^*|\xi)}\right] \quad (6)$$

where $\xi^*$ is the proposed value, $\xi$ is the current value, and $Q(\xi^*|\xi)$ is the probability of proposing $\xi^*$ given $\xi$. Making the dependency on $K$ in our finite model explicit and defining $P(K)$ to be the prior probability of $K$, we can factorize $P(\mathbf{X}, \xi)$ into the known probabilities $P(\mathbf{X}|\mathbf{Z},\mathbf{Y})P(\mathbf{Y}|K)P(\mathbf{Z}|K)P(K)$.

Using this proposal, a hidden cause is added with probability $K_+/K$. An empty column is added to $\mathbf{Z}$, and a corresponding row of $\mathbf{Y}$ is generated by sampling according to Eqn. 4. The probability of proposing this new configuration of $K$, $\mathbf{Z}$, and $\mathbf{Y}$ is thus $(K_+/K)p^{\sum_t y_{k,t}}(1-p)^{\sum_t (1-y_{k,t})}$ where $k$ is the index of the new column. To return to the previous configuration we may delete any hidden cause with the same value as our proposed new row of $\mathbf{Y}$. The probability of choosing such a row is $\delta/(K+1)$ where $\delta$ is the number of rows of $\mathbf{Y}$ (including the new row) which are identical to the proposed new row. Consequently, the ratio of proposal probabilities is

$$\frac{Q(\xi|\xi^*)}{Q(\xi^*|\xi)} = \frac{\frac{\delta}{K+1}}{\frac{K_+}{K}p^{\sum_t y_{k,t}}(1-p)^{\sum_t (1-y_{k,t})}}. \quad (7)$$

The ratio of the probabilities of the resulting configurations needs to take into account the difference in the probability of $\mathbf{Y}$, which is just the probability of the new row of $\mathbf{Y}$, $p^{\sum_t y_{k,t}}(1-p)^{\sum_t (1-y_{k,t})}$, the different probabilities of $\mathbf{Z}$ with and without the new column (with the corresponding changes in $K$), and the different probabilities of $K$. This gives the ratio

$$\frac{P(\mathbf{X}, \xi^*)}{P(\mathbf{X}, \xi)} = \frac{p^{\sum_t y_{k,t}}(1-p)^{\sum_t (1-y_{k,t})} P(\mathbf{Z}|K+1)P(K+1)}{P(\mathbf{Z}|K)P(K)}$$

Putting together this with Eqn. 7 gives

$$A(\xi^*, \xi) = \min\left[1, \frac{\frac{\delta}{K+1} P(\mathbf{Z}|K+1)P(K+1)}{\frac{K_+}{K} P(\mathbf{Z}|K)P(K)}\right]. \quad (8)$$

A similar argument yields the acceptance probability for the proposal to delete a hidden cause with no links.

$$A(\xi^*,\xi) = \min\left[1, \frac{\frac{K_+}{K-1}P(\mathbf{Z}|K-1)P(K-1)}{\frac{\delta}{K}P(\mathbf{Z}|K)P(K)}\right]. \quad (9)$$

To complete the specification of the algorithm, we need a scheme for sampling $\mathbf{Z}$ and $\mathbf{Y}$. We use Gibbs sampling, drawing each component of the two matrices from the distributions $P(z_{i,k}|\mathbf{X},\mathbf{Z}_{-i,k},\mathbf{Y})$ and $P(y_{k,t}|\mathbf{X},\mathbf{Z},\mathbf{Y}_{-k,t})$, where $\mathbf{Z}_{-i,k}$ is all values of $\mathbf{Z}$ except for $z_{i,k}$ and $\mathbf{Y}_{-k,t}$ is all of the values of the matrix $\mathbf{Y}$ except for $y_{k,t}$. As both $z_{i,k}$ and $y_{k,t}$ are binary, these probabilities can be computed by enumeration. From our generative model and Bayes' rule we have

$$\begin{aligned}&P(z_{i,k}=a|\mathbf{X},\mathbf{Z}_{-i,k},\mathbf{Y})\\&\propto P(\mathbf{X}|\mathbf{Z}_{-i,k},\mathbf{Y},z_{i,k}=a)P(z_{i,k}=a|\mathbf{z}_{-i,k})\end{aligned}$$

where $P(\mathbf{X}|\mathbf{Z}_{-i,k},\mathbf{Y},z_{i,k}=a)$ is specified by Equation 3, and $P(z_{i,k}=a|\mathbf{z}_{-i,k})$ results from our prior on $\mathbf{Z}$. It follows from Eqn. 5 that

$$P(z_{i,k}=1|\mathbf{z}_{-i,k}) = \bar{\theta}_k = \frac{m_{-i,k}+\frac{\alpha}{K}}{N} \quad (10)$$

where $\mathbf{z}_{-i,k}$ is all of the entries of column $k$ of $\mathbf{Z}$ except row $i$ and $m_{-i,k}=\sum_{j\neq i}^N z_{j,k}$. Consequently, we obtain

$$\begin{aligned}&P(z_{i,k}=a|\mathbf{X},\mathbf{Z}_{-i,k},\mathbf{Y}) \quad (11)\\&\propto \bar{\theta}_k^a(1-\bar{\theta}_k)^{(1-a)}\prod_{t=1}^T(1-(1-\lambda)^{\mathbf{z}_{i,:}\cdot\mathbf{Y}_{:,t}}(1-\epsilon))|_{z_{i,k}=a}\end{aligned}$$

where $|_{z_{i,k}=a}$ means to replace $z_{i,k}$ with $a$ in the preceding expression. Proceeding similarly for $y_{k,t}$ gives

$$\begin{aligned}&P(y_{k,t}=a|\mathbf{Z},\mathbf{X},\mathbf{Y}_{-k,t}) \quad (12)\\&\propto p^a(1-p)^{1-a}\prod_{i=1}^N(1-(1-\lambda)^{\mathbf{z}_{i,:}\cdot\mathbf{Y}_{:,t}}(1-\epsilon))|_{y_{k,t}=a}.\end{aligned}$$

This yields the RJMCMC sampler in Algorithm 1.

### 4.2 Gibbs Sampler for the Infinite Case

When $K \to \infty$, we no longer require dimension-jumping moves, and can simply use a Gibbs sampler to infer $\mathbf{Y}$ and $\mathbf{Z}$. The only difference from the Gibbs component of the RJMCMC algorithm outlined above is the scheme for sampling $z_{i,k}$. We can find $P(z_{i,k}=1|\mathbf{z}_{-i,k})$ by exploiting the exchangeability of the IBP. Since any ordering of the customers results in the same distribution, we can assume that the $i^{\text{th}}$ customer is the last to enter the restaurant. Accordingly, they should sample each dish that has previously been tasted by $m_{-i,k}$ customers with probability $\frac{m_{-i,k}}{N}$, and try a Poisson($\frac{\alpha}{N}$) number of new dishes, for which $m_{-i,k}=0$. Thus, we need to consider two cases in sampling $z_{i,k}$ in our Gibbs sampler: the case where $m_{-i,k}>0$, and the case where $m_{-i,k}=0$.

**Algorithm 1** RJMCMC sampler for hidden causes

1: **for** $r=1,\ldots,$ number of iterations **do**
2:   **for** $i=1,\ldots,N$ **do**
3:     randomly select column $k$ of $\mathbf{Z}$
4:     **if** $m_{i,k}>0$ **then**
5:       propose adding a new cause
6:       accept according to Eqn. 8
7:     **else**
8:       propose deleting this unlinked cause
9:       accept according to Eqn. 9
10:     **end if**
11:     **for** $k=1,\ldots,K$ **do**
12:       sample $z_{i,k}$ according to Eqn. 11
13:     **end for**
14:     **for all** $y_{k,t}\in\mathbf{Y}$ **do**
15:       sample $y_{k,t}$ according to Eqn. 12
16:     **end for**
17:   **end for**
18: **end for**

When $m_{-i,k}>0$, $P(z_{i,k}=1|\mathbf{z}_{-i,k})$ is essentially the same as in the finite model, being given by Eqn. 10 with $\bar{\theta}_k = \frac{m_{-i,k}}{N}$. The case where $m_{-i,k}=0$ requires more careful treatment. In our non-parametric approach, $\mathbf{Z}$ is a matrix with infinitely many columns (and $\mathbf{Y}$ is a matrix with infinitely many rows). In practice, only the non-zero columns of the matrix can be held in memory, but we still need to sample those columns. To do so let $K_i^{\text{new}}$ be the number of columns of $\mathbf{Z}$ which contain a 1 only in row $i$. Then we have

$$\begin{aligned}&P(K_i^{\text{new}}|\mathbf{X}_{i,1:T},\mathbf{Z}_{i,1:K+K_i^{\text{new}}},\mathbf{Y}) \quad (13)\\&\propto P(\mathbf{X}_{i,1:T}|\mathbf{Z}_{i,1:K+K_i^{\text{new}}},\mathbf{Y},K_i^{\text{new}})P(K_i^{\text{new}})\end{aligned}$$

where the prior $P(K_i^{\text{new}})$ is Poisson($\frac{\alpha}{N}$), and we find $P(\mathbf{X}_{i,1:T}|\mathbf{Z}_{i,1:K+K_i^{\text{new}}},\mathbf{Y},K_i^{\text{new}})$ by marginalizing over $\mathbf{Y}^{\text{new}}=\mathbf{Y}_{K+1:K+K_i^{\text{new}},t}$, the new rows of $\mathbf{Y}$. As each entry of $\mathbf{X}$ is independent, we have

$$\begin{aligned}&P(\mathbf{X}_{i,1:T}|\mathbf{Z}_{i,1:K+K_i^{\text{new}}},\mathbf{Y},K_i^{\text{new}}) \quad (14)\\&= \prod_{t=1:T}P(x_{i,t}|\mathbf{Z}_{i,1:K+K_i^{\text{new}}},\mathbf{Y},K_i^{\text{new}})\end{aligned}$$

marginalizing over $\mathbf{Y}^{\text{new}}$ gives

$$\begin{aligned}&P(x_{i,t}=1|\mathbf{Z}^{\text{new}},\mathbf{Y},K_i^{\text{new}})\\&= \sum_{\mathbf{Y}^{\text{new}}}P(x_{it}=1|\mathbf{Z}^{\text{new}},\mathbf{Y}^{\text{new}})P(\mathbf{Y}^{\text{new}})\\&= \sum_{m=0}^{K_i^{\text{new}}}[1-\eta(1-\lambda)^m(1-\epsilon)]p^m(1-p)^{K_i^{\text{new}}-m}\binom{K_i^{\text{new}}}{m}\\&= 1-(1-\epsilon)\eta\sum_{m=0}^{K_i^{\text{new}}}[(1-\lambda)p]^m(1-p)^{K_i^{\text{new}}-m}\binom{K_i^{\text{new}}}{m}\\&= 1-(1-\epsilon)\eta(1-\lambda p)^{K_i^{\text{new}}} \quad (15)\end{aligned}$$

where $\eta=(1-\lambda)^{\mathbf{z}_{i,1:K}\cdot\mathbf{Y}_{1:K,t}}$. The last step makes use of the binomial theorem. This gives us all we need to compute the conditional distribution over $K_i^{\text{new}}$ defined in Eqn. 13. In theory, sampling from this distribution would require evaluating it at all possible values of $K_i^{\text{new}}$. We approximate this by sampling from the distribution over $K_i^{\text{new}} \leq 10$.

**Algorithm 2** Gibbs sampler for hidden causes
1: **for** $r = 1, \ldots,$ number of iterations **do**
2:   **for** $i = 1, \ldots, N$ **do**
3:     **for** $k = 1, \ldots, K$ **do**
4:       **if** $m_{-i,k} > 0$ **then**
5:         sample $z_{i,k}$ according to Eqn. 11
6:       **else**
7:         mark $z_{i,k}$ to be zeroed
8:       **end if**
9:     **end for**
10:     zero marked $z_{i,k}$'s
11:     sample $K_i^{\text{new}}$ according to Eqn. 13
12:     $\mathbf{Z}_{i,K+1:K+K_i^{\text{new}}} \leftarrow 1$
13:     **for all** $y_{j,t} \in \mathbf{Y}_{K+1:K+K_i^{\text{new}}, 1:T}$ **do**
14:       sample $y_{j,t}$ according to Eqn. 12
15:     **end for**
16:     $K \leftarrow K + K_i^{\text{new}}$
17:   **end for**
18:   **for all** $y_{k,t} \in \mathbf{Y}$ **do**
19:     sample $y_{k,t}$ according to Eqn. 12
20:   **end for**
21:   remove columns with $\sum_i z_{i,k} = 0$ from $\mathbf{Z}$
22:   remove corresponding rows from $\mathbf{Y}$
23: **end for**

While $\mathbf{Y}$ technically has an infinite number of rows, in practice we need only sample $y_{k,t}$ for those rows that correspond to non-zero columns of $\mathbf{Z}$. The pseudocode for our Gibbs sampler is given in Algorithm 2.

## 5 Evaluation with Synthetic Data

We evaluated the RJMCMC algorithm for the finite model and the Gibbs sampler for the infinite model on two tasks using simulated data. First, we examined the ability of both algorithms to recover the true number of hidden causes used to generate a dataset. The data were generated by fixing the number of observations, $N$, and varying the number of hidden causes, $K$. For each value of $K$, 10 different datasets were generated by using rejection sampling to draw a matrix $\mathbf{Z}$ of the appropriate dimensionality from the IBP, drawing $\mathbf{Y}$ according to Eqn. 4, and then drawing $\mathbf{X}$ according to Eqn. 3. RJMCMC and Gibbs were both initialized with either an empty $\mathbf{Z}$ matrix ($K = 1$ for RJMCMC) or random $\mathbf{Z}$ and $\mathbf{Y}$ matrices with $K = K_+ = 10$, and then run for 500 iterations on each dataset. The other model parameters were fixed at $T = 500$, $\alpha = 3$, $\epsilon = 0.01$, $\lambda = 0.9$, and $p = 0.1$.

Our results are shown in Fig. 3. The Gibbs sampler slightly over-estimates the number of hidden causes, but generally produces results that are close to the true dimensionality regardless of initialization. In contrast, RJMCMC appears to be affected by initialization. In particular, it systematically under-estimates the true dimensionality when initialized with $K = 1$. This is a result of poor mixing: while proposals to add hidden causes are often accepted, the new values of $\mathbf{Y}$ associated with those causes are typically inconsistent with the structure of $\mathbf{X}$, and consequently the new causes do not obtain links to observable nodes. The new causes are thus quickly deleted. A new cause might be generated with an appropriate set of $\mathbf{Y}$ values given sufficiently many sampling iterations (see Fig. 4), but in a short run like that used here, slow mixing results in a strong influence of initialization.

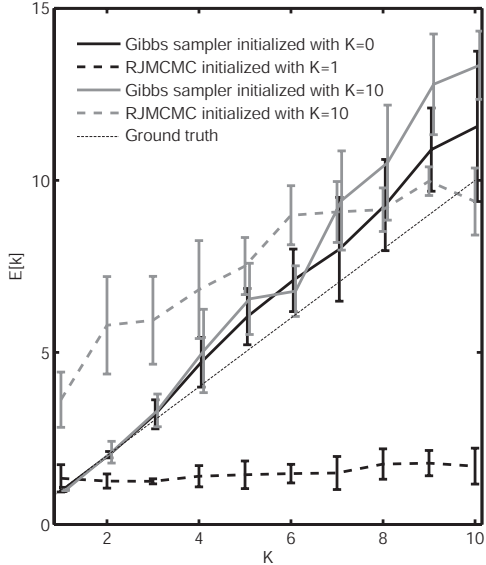

Figure 3: Learning the number of hidden causes using both RJMCMC and Gibbs sampling. Each line show the mean and standard deviation of the expected value of the dimensionality of the model ($K$ for RJMCMC, and $K_+$ for Gibbs) taken over 500 iterations of sampling for each of 10 datasets.

Our second evaluation compared the ability of the two algorithms to recover specific structures. We manually specified four $\mathbf{Z}$ matrices, and then generated 10 datasets for each using the procedure outlined above with $T = 150$ $\alpha = 3$, $\epsilon = 0.01$, $\lambda = 0.9$, and $p = 0.1$. Both RJMCMC and Gibbs were initialized with an empty $\mathbf{Z}$ matrix ($K = 1$ for RJMCMC). We used two measures to evaluate performance. First, *in-degree error*, which we define to be the difference between the true in-degree and the expected in-degree of the observed nodes computed over samples. This is computed by taking the sum absolute difference between $\text{diag}(\mathbf{ZZ}^T)$, the in-degree of the observable nodes, and $\text{diag}(E[\mathbf{ZZ}^T])$, the expected in-degree computed over the samples. And second, the *structure error*, which we define to be the sum absolute difference between the upper triangular portion of $\mathbf{ZZ}^T$ and $E[\mathbf{ZZ}^T]$. Each element of the upper triangular portion of $\mathbf{ZZ}^T$ is a count of the number of hidden causes shared by a pair of observable variables, the sum difference is a general measure of graph dissimilarity.

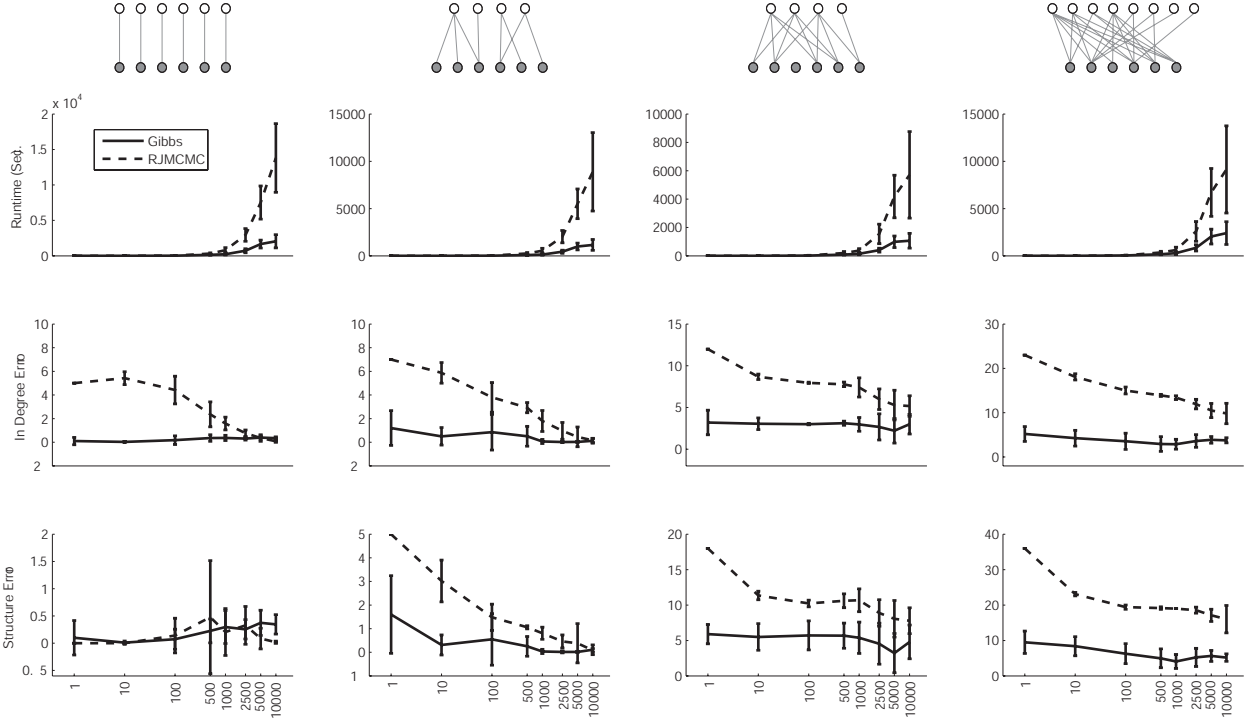

Figure 4: Recovering causal structure with RJMCMC and Gibbs sampling. From left to right the columns are results for a degree 1 bipartite graph ($K = 6$), disconnected graph ($K = 4$), an "undercomplete" random graph with fewer causes than observations ($K = 4$), and an "overcomplete" random graph with more causes than observations ($K = 8$). The top row shows the true structures, the second row shows mean runtime in wall clock seconds as a function of the number of iterations of sampling. The third and fourth rows show the in-degree error and structure error for each algorithm, as defined in the main text, on the same axis of number of iterations. Error bars are symmetric, and indicate one standard deviation over 10 datasets.

The results are shown in Fig. 4. The Gibbs sampler consistently recovers a structure close to the truth, and does so in surprisingly few iterations. This reflects a tendency to move quickly to a good solution, and then minimally explore the space around that solution. The variance of the results grow slightly with more iterations, reflecting greater exploration of the space of structures. In contrast, RJMCMC performs poorly for all but the largest number of iterations. This is a reflection of the fact that it mixes slowly, taking a long time to increase the dimension of a model. It should be noted that the poor performance on the overcomplete graph is not so much a problem with the algorithms as an indication of an unavoidable problem with identifiability in overcomplete models. For instance, in this particular graph there is no information, short of the prior on $\mathbf{Y}$, that can be used to distinguish causal nodes 5 and 6 from a hypothetical single combined node.

## 6 Inferring Stroke Localizations

We used a subset of the Mount Sinai Stroke Data Bank [13] to illustrate our approach to inferring hidden causes with real data. This data bank consists of stroke signs exhibited by patients admitted to an acute stroke unit at Mount Sinai Hospital, together with lesion localization evaluations made by neurologist with special stroke expertise (the data were collected from a standardized neurological assessment including a detailed neurologic examination). In the language of the preceding sections, the signs are our observed variables, and the localizations are our hidden causes. The raw data bank consisted of 38 signs and 14 localizations. Some signs were left-right variables and some were graded in degrees of severity. For each patient, signs were binarized in two steps. First, graded signs like *decreased level of consciousness* and *comprehension deficit severity* were assigned a 1 if any level was indicated at all and 0 if no indication was made. Second, for "sided" signs like *visual field deficit* and *abnormal deep tendon reflex* we created two variables, one each for left and right, and assigned a 1 to the vari-able corresponding to the side on which the sign was observed. Of the resulting 56 sign variables, only 42 were expressed by at least one patient. The mean number of signs per patient was 8.24 and the mean number of stroke localizations was 1.96. Every localization was found in at least one patient, although there were five localizations that were found in only one patient.

Although the ground truth localizations were known, we inferred the localizations, $\mathbf{Y}$, and their causal relationships to the signs, $\mathbf{Z}$, directly from the signs exhibited by the patients, $\mathbf{X}$, using our Gibbs sampler. In addition, we placed a Beta$(1,1)$ prior on $\lambda$, $\epsilon$, and $p$, and a Gamma$(1,1)$ prior on $\alpha$, and sampled them as well using Metropolis updates for $\lambda$ and $\epsilon$ and Gibbs for $p$ and $\alpha$. These hyperparameters have interpretations: $p$ measures of the incidence rate of localizations (the approximate ground truth for the data bank is $p = 0.14$), $\epsilon$ is the rate of spontaneous sign expression (noise), $\lambda$ is a measure of how reliably a localization (or combination of localizations) gives rise to a sign, and $\alpha$ and $K$ together measure of the number of hidden localizations (ground truth is $K = 14$).

Trace plots for the hyperparameters over 20,000 iterations of Gibbs sampling appear in Fig. 5. Interpretation of these results should be considered under the caveat that there are few datapoints in this data bank, and that all the data is stroke-specific. The posterior distribution on $\epsilon$ favored low values, suggesting that the prevalence of these signs in the absence of a particular stroke localization in these patients is low. That values of $\lambda$ are reasonably high reflects the fact that the localizations responsible for producing a particular sign produce that sign with high probability. The posterior distributions on $p$ and $K$ favored values slightly higher and lower than the ground truth, respectively, but these parameters should be expected to be coupled, since they influence the overall prevalence of signs. The under-estimate of $K$ is not unexpected, due to the paucity of data for many localizations. In fact, there were only nine localizations exhibited by at least two patients, providing a closer correspondence to the values favored by the sampler.

The causal structure with the highest posterior probability in our set of samples is shown in Fig. 6. We can attempt to interpret the hidden causes by examining the signs to which they are connected. We showed the clusters of signs corresponding to the hidden causes found by the algorithm to a clinical neurologist familiar with the domain, who concluded that the localizations were somewhat general but not inappropriately confounded. These observations came with the caveat that the loss of degree information and the abridgement of the typical clinical sign and localization domain made precise localizations difficult. As a further encouraging sign, we note that the data bank contained no bilateral stroke sufferers, and the recovered graph reflects this by correctly separating signs that are caused by infarcts in each hemisphere.

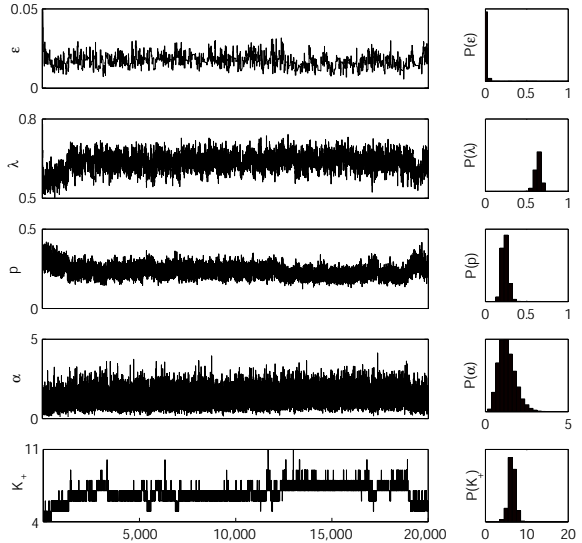

Figure 5: Trace plots and histograms for the Gibbs sampler applied to the signs exhibited by 50 stroke patients. The left column shows the current value of $\epsilon, \lambda, p, \alpha$, and $K_+$ as the sampler progressed, where $K_+$ is obtained by examining the current $\mathbf{Z}$ sample. The right column shows histograms of the same variables computed over the samples.

## 7 Summary and Conclusion

In this paper we developed and demonstrated a nonparametric Bayesian technique for simultaneously inferring the existence and connectivity of hidden causes. Our algorithm correctly recovers the number of hidden causes that influence a set of observed variables, and can be used to obtain reasonably good estimates of the causal structure underlying a domain. This approach provides a promising foundation for the development of Bayesian models that can be used to learn both the structure relating a set of observed variables, and the hidden causes that influence those variables. In particular, our algorithm can easily be integrated with existing MCMC methods for Bayesian structure learning (e.g., [1]).

Our results suggest a number of future directions. First, while our focus here was on the case where $p$ and $\lambda$ were shared by all hidden causes, variations on the algorithm we describe could be applied in the case where these hyperparameters vary across causes, extending the applicability of the model. Second, the slow mixing exhibited by RJMCMC requires further investigation. While birth/death proposals of the kind we used here are common, it may be possible to develop a faster-mixing proposal by drawing inspiration from the Gibbs sampler for our non-parametric model, adding and deleting nodes with a single link attached.

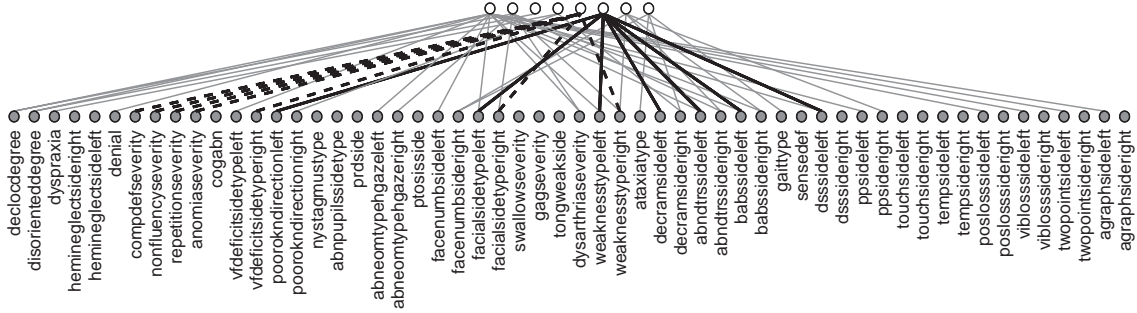

Figure 6: Causal structure with highest posterior probability. Two grouping of signs are highlighted. In solid black, we find a grouping of poor optokinetic nystagmus, lack of facial control, weakness, decreased rapid alternating movements, abnormal deep tendon reflexes, Babinski sign, and double simultaneous stimulation neglect, all on the left side, consistent with a right frontal/parietal infarct. In dashed black, we find a grouping of comprehension deficit, non-fluency, repetition, anomia, visual field deficit, facial weakness, and general weakness, with the latter three on the right side, generally consistent, in part, with a left temporal infarct.

The non-parametric Bayesian methods explored in this paper make it possible to learn Bayesian networks with infinitely many nodes. While this might seem intractable at first glance, assuming that the number of nodes is unbounded actually removes the formal problems involved in inferring hidden causes, and leads to a simple algorithm with broad applicability.

**Acknowledgments**

This work was supported by NIH-NINDS R01 NS 50967-01 as part of the NSF/NIH Collaborative Research in Computational Neuroscience Program. We thank Leigh R. Hochberg for graciously sharing his expertise, Michael J. Black for suggestions, and Stanley Tuhrim of Mt. Sinai Hospital for data and permissions.

## References

[1] N. Friedman and D. Koller, "Being Bayesian about network structure," *UAI 16*, 2000.

[2] D. Heckerman, "A tutorial on learning with Bayesian networks," in *Learning in Graphical Models*, M. I. Jordan (ed.). MIT Press, 1998.

[3] M. Shwe, B. Middleton, D. Heckerman, M. Henrion, E. Horvitz, H. Lehmann, and G. Cooper, "Probabilistic diagnosis using a reformulation of the INTERNIST-1/QMR knowledge base I. the probabilistic model and inference algorithms," *Methods of Information in Medicine*, vol. 30, pp. 241–255, 1991.

[4] G. Elidan and N. Friedman, "The information bottleneck expectation maximization algorithm," *UAI 19*, 2003.

[5] P. Green, "Reversible jump Markov chain Monte Carlo computation and Bayesian model determination," *Biometrika*, pp. 711–732, 1995.

[6] A. C. Courville, N. D. Daw, and D. S. Touretzky, "Similarity and discrimination in classical conditioning: A latent variable account," *NIPS 17*, 2005.

[7] G. Orban, J. Fiser, R. N. Aslin, and M. Lengyel, "Bayesian model learning in human visual perception," *NIPS 18*, 2006.

[8] C. Antoniak, "Mixtures of Dirichlet processes with applications to Bayesian nonparametric problems," *The Annals of Statistics*, vol. 2, pp. 1152–1174, 1974.

[9] R. M. Neal, "Markov chain sampling methods for Dirichlet process mixture models," *Journal of Computational and Graphical Statistics*, vol. 9, pp. 249–265, 2000.

[10] C. Rasmussen, "The infinite Gaussian mixture model," *NIPS 12*, 2000.

[11] T. L. Griffiths and Z. Ghahramani, "Infinite latent feature models and the Indian buffet process," Gatsby Unit, Tech. Rep. 2005-001, 2005.

[12] J. Pearl, *Probabilistic reasoning in intelligent systems*, Morgan Kauffman, San Francisco, 1988.

[13] S. Tuhrim, J. Reggia, and S. Goodall, "An experimental study of criteria for hypothesis plausibility," *Journal of Experimental and Theoretical Artificial Intelligence*, vol. 3, pp. 129–144, 1991.